\documentclass{article}

    \PassOptionsToPackage{numbers, compress}{natbib}

    \usepackage[preprint]{neurips_2020}

\usepackage{amsmath}
\usepackage{times}
\usepackage{url}
\usepackage{amssymb}
\usepackage{xcolor}
\usepackage{latexsym}
\usepackage{subfig}
\usepackage{varwidth}
\usepackage{graphicx}
\usepackage{cases}
\usepackage{placeins}
\usepackage{array, boldline, makecell, booktabs}
\usepackage{multirow}
\usepackage{multicol}
\usepackage{latexsym}
\usepackage{graphicx}
\usepackage{placeins}
\usepackage{booktabs}
\definecolor{agr}{rgb}{0.0, 0.5, 0.0}
\newcommand{\comment}[1]{}
\makeatletter
\def\thickhline{%
  \noalign{\ifnum0=`}\fi\hrule \@height \thickarrayrulewidth \futurelet
   \reserved@a\@xthickhline}
\def\@xthickhline{\ifx\reserved@a\thickhline
               \vskip\doublerulesep
               \vskip-\thickarrayrulewidth
             \fi
      \ifnum0=`{\fi}}
\makeatother
\usepackage{xcolor,colortbl}
\newlength{\thickarrayrulewidth}
\definecolor{Gray}{gray}{0.85}
\definecolor{Blue}{HTML}{cceeff}
\definecolor{Blued}{HTML}{66ccff}
\usepackage{subfig}
\usepackage{bm}

\usepackage[utf8]{inputenc} 
\usepackage[T1]{fontenc}    
\usepackage{hyperref}       
\usepackage{url}            
\usepackage{booktabs}       
\usepackage{amsfonts}       
\usepackage{nicefrac}       
\usepackage{microtype}      

\definecolor{blue_c}{RGB}{40, 116, 166}
\definecolor{orange_c}{RGB}{175, 96, 26 }

\newcolumntype{?}{!{\vrule width 3\arrayrulewidth}}

\title{Bi-ISCA: Bidirectional Inter-Sentence Contextual Attention Mechanism for Detecting Sarcasm in User Generated Noisy Short Text}

\author{%
  Prakamya Mishra \\
  Shiv Nadar University\\
  \texttt{pm669@snu.edu.in} \\
  \And
  Saroj Kaushik \\
  Shiv Nadar University\\
  \texttt{saroj.kaushik@snu.edu.in} \\
  \And
  Kuntal Dey \\
  Accenture Technology Labs\\
  \texttt{kuntal.dey@accenture.com} \\
}

\begin{document}

\maketitle

\begin{abstract}
  Many online comments on social media platforms are hateful, humorous, or sarcastic. The sarcastic nature of these comments (especially the short ones) alters their actual implied sentiments, which leads to misinterpretations by the existing sentiment analysis models. A lot of research has already been done to detect sarcasm in the text using user-based, topical, and conversational information but not much work has been done to use inter-sentence contextual information for detecting the same. This paper proposes a new state-of-the-art deep learning architecture that uses a novel Bidirectional Inter-Sentence Contextual Attention mechanism (Bi-ISCA) to capture inter-sentence dependencies for detecting sarcasm in the user-generated short text using only the conversational context. The proposed deep learning model demonstrates the capability to capture explicit, implicit, and contextual incongruous words \& phrases responsible for invoking sarcasm. Bi-ISCA generates state-of-the-art results on two widely used benchmark datasets for the sarcasm detection task (Reddit and Twitter). To the best of our knowledge, none of the existing state-of-the-art models use an inter-sentence contextual attention mechanism to detect sarcasm in the user-generated short text using only conversational context.
\end{abstract}

\section{Introduction} \label{introduction}
Sentiment analysis is one of the most important natural language processing (NLP) applications. Its goal is to identify, extract, quantify, and study subjective information. The sudden rise in the usage of social media platforms as a means of communication has led to a vast amount of data being shared between its users on a wide range of topics. This type of data is very helpful to several organizations for analyzing the sentiments of people towards products, movies, political events, etc. Understanding the unique intricacies of the human language remains one of the most important pending NLP problems of this time. Humans regularly use sarcasm as a crucial part of the day-to-day conversations when venting, arguing, or maybe engaging on social media platforms. Sarcastic remarks on these platforms inflict problems on the existing sentiment analysis systems in identifying the true intentions of the users.

The Cambridge Dictionary\footnote{https://dictionary.cambridge.org/} describes sarcasm as an irony conveyed hilariously or amusingly to criticize something. Sarcasm may not show criticism on the surface but instead might have a criticizing implied meaning. Such a figurative aspect of sarcasm makes it difficult to be detected in the modern micro texts \cite{ghosh-veale-2016-fracking}. Several linguistic research has been done to analyze different aspects of sarcasm. Kind of responses evoked because of comments has been considered a major indicator of sarcasm \cite{EISTERHOLD20061239}. \citet{WILSON20061722} states that circumstantial incongruity between a comment and its corresponding contextual information plays an important role in implying sarcasm.

Previous research works have used policy-based, statistical, and deep-learning-based methods for detecting sarcasm. The use of contextual information like conversational context, author personality features, or prior knowledge of the topic, have proved to be very useful. \citet{khattri2015your} used sentiments of the author's historical tweets as context. \citet{10.1145/2684822.2685316} used personality features like the author's familiarity with twitter, language (structure and word usage), and the author's familiarity with sarcasm (history of previous sarcastic tweets) for consolidating context. \citet{bamman2015contextualized} explored the use of historical terms, topics, and sentiments along with profile information as the author's context. They also exploited the use of conversational context like the immediate previous tweets in the thread. \citet{joshi-etal-2015-harnessing} demonstrated that concatenation of preceding comment with the objective comment in a discussion forum led to an increase in the precision score.

Overall in recent years a lot of work has been done to use different types of contextual information for sarcasm detection but none of them have used inter-sentence dependencies. In this paper, we propose a novel Bidirectional Inter-Sentence Contextual Attention mechanism (Bi-ISCA) based deep learning neural network for sarcasm detection. The main contribution of this paper can be summarised as follows:
\begin{itemize}
    \item We propose a new state-of-the-art deep learning architecture that uses a novel Bidirectional Inter-Sentence Contextual attention mechanism (Bi-ISCA) for detecting sarcasm in short texts (short texts are more difficult to analyze due to shortage of contextual information). 
    \item Bi-ISCA focuses on only using the conversational contextual comment/tweet for detecting sarcasm rather than using any other topical/personality-based features, as using only the contextual information enriches the model's ability to capture syntactical and semantical textual properties responsible for invoking sarcasm.
    \item We also explain model behavior and predictions by visualizing attention maps generated by Bi-ISCA, which helps in identifying significant parts of the sentences responsible for invoking sarcasm.
\end{itemize}

The rest of the paper is organized as follows. Section \ref{relatedwork} describes the related work. Then section \ref{model}, explains the proposed model architecture for detecting sarcasm. Section \ref{evaluation} will describe the datasets used, pre-processing pipeline, and training details for reproducibility. Then experimental results are explained in section \ref{result} and section \ref{discussion} illustrates model behavior and predictions by visualizing attention maps. Finally we conclude in section \ref{conclude}.

\section{Related Work} \label{relatedwork}
A diverse spectrum of approaches has been used to detect sarcasm. Recent sarcasm detection approaches have either mainly focused on using machine learning based approaches that leverage the use of explicitly declared relevant features or they focus on using neural network based deep learning approaches that do not require handcrafted features. Also, the recent advances in using deep learning for preforming natural language processing tasks have led to a promising increase in the performance of these sarcasm detection systems.

A lot of research has been done using bag of words as features. However, to improve performance, scholars started to explore the use of several other semantic and syntactical features like punctuations \cite{tsur2010icwsm}; emotion marks and intensifiers \cite{liebrecht-etal-2013-perfect}; positive verbs and negative phrases \cite{DBLP:conf/emnlp/RiloffQSSGH13}; polarity skip grams \cite{reyes2013multidimensional}; synonyms \& ambiguity\cite{barbieri-etal-2014-modelling}; implicit and explicit incongruity-based \cite{joshi-etal-2015-harnessing}; sentiment flips \cite{10.1145/2684822.2685316}; affect-based features derived from multiple emotion lexicons \cite{10.1145/2930663}.

Every day an enormous amount of short text data is generated by users on popular social media platforms like Twitter\footnote{www.twitter.com/} and Reddit\footnote{www.reddit.com/}. Easy accessibility of such data sources has enticed researchers to use them for extracting user-based and discourse-based features. \citet{hazarika-etal-2018-cascade} utilized contextual information by making user-embeddings for capturing indicative behavioral traits. These user-embeddings incorporated personality features along with the author's writing style (using historical posts). They also used discourse comments along with background cues and topical information for detecting sarcasm. They performed their experiments on the largest Reddit dataset SARC \cite{khodak2018corpus}. Many have only used the target text for classification purposes, where a target text is a textual unit that has to be classified as sarcastic or not. Simply using gated recurrent units (GRU) \cite{DBLP:journals/corr/ChoMGBSB14} or long short term memory (LSTM) \cite{doi:10.1162/neco.1997.9.8.1735} do not capture in between interactions of word pairs which makes it difficult to model contrast and incongruity. \citet{tay-etal-2018-reasoning} were able to solve this problem by looking in-between word pairs using a multi-dimensional intra-attention recurrent network. They focused on modeling the intra-sentence relationships among the words. \citet{8949523} exploited the use of a multi-head attention mechanism \cite{NIPS2017_7181} which could capture dependencies between different representations subspaces in different positions. Their model consisted of a word encoder for generating new word representations by summarizing comment contextual information in a bidirectional manner. On top of that, they used multi-head attention for focusing on different contexts of a sentence, and in the end, a simple multi-layer perceptron was used for classification.

There has not been much work done in conversation dependent (comment and reply) approaches for sarcasm detection. \citet{DBLP:journals/corr/abs-1809-03051} proposed a model that not only used information from the target utterance but also used its conversational context to perceive sarcasm. They aimed to detect sarcasm by just using the sequences of sentences, without any extra knowledge about the user and topic. They combined the predictions from utterance-only and conversation-dependent parts for generating its final prediction which was able to capture the words responsible for delivering sarcasm. \citet{DBLP:journals/corr/GhoshFM17} also modeled conversational context for sarcasm detection. They also attempted to derive what parts of the conversational context triggered a sarcastic reply. Their proposed model used sentence embeddings created by taking an average of word embeddings and a sentence-level attention mechanism was used to generate attention induced representations of both the context and the response which was later concatenated and used for classification. 

Among all the previous works, \cite{DBLP:journals/corr/abs-1809-03051} and \cite{DBLP:journals/corr/GhoshFM17} share similar motives of detecting sarcasm using only the conversational context. However, we introduce a novel Bidirectional Inter-Sentence Contextual Attention mechanism (Bi-ISCA) for detecting sarcasm. Unlike previous works, our work considers short texts for detecting sarcasm, which is far more challenging to detect when compared to long texts as long texts provide much more contextual information.

\begin{figure}[t] 
    \centering
    \includegraphics[width=0.85\textwidth]{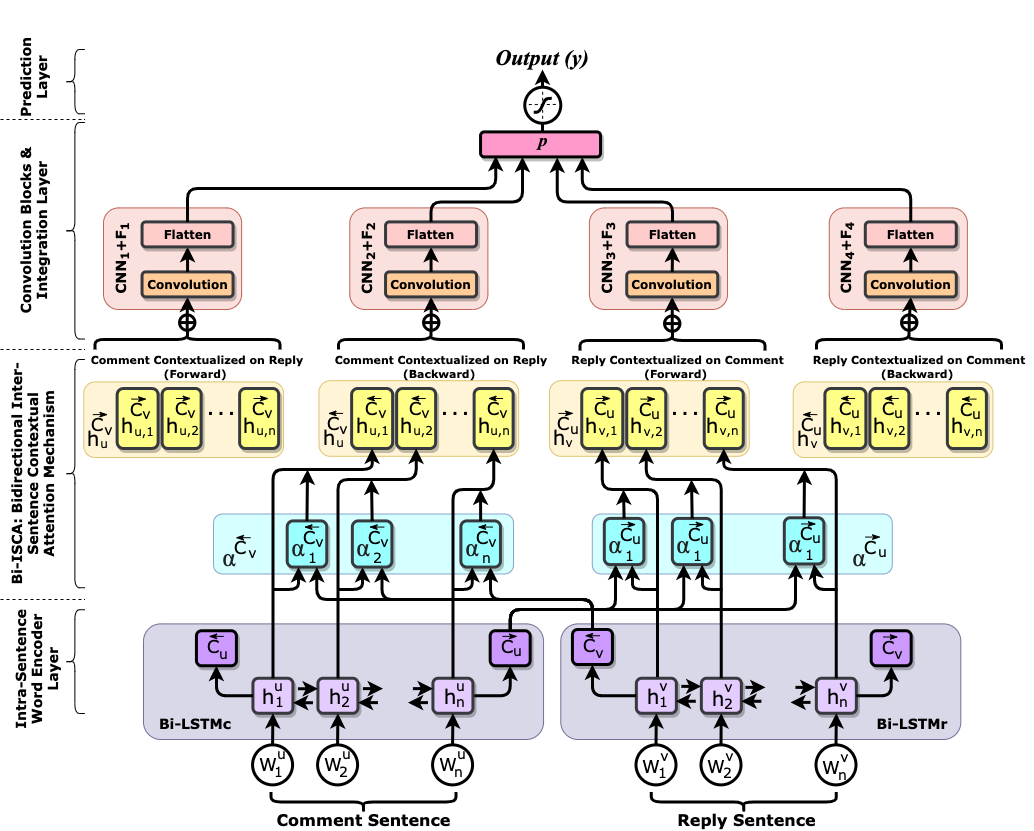}
    \caption{Bi-ISCA: Bi-Directional Inter-Sentence Contextual Attention Mechanism for Sarcasm Detection.}
    \label{fig:main_model}
\end{figure}

\section{Model} \label{model}

This section will introduce the proposed Bi-ISCA: Bidirectional Inter Sentence Contextual Attention based neural network for sarcasm detection (as shown in Figure \ref{fig:main_model}). Sarcasm detection is a binary classification task that tries to predict whether a given comment is sarcastic or not. The proposed model uses comment-reply pairs for detecting sarcasm. The input to the model is represented by \(U = [W_1^u,W_2^u,....,W_n^u]\) and \(V = [W_1^v,W_2^v,....,W_n^v]\), where U represents the comment sentence and V represents the reply sentence (both sentences padded to a length of n). Here, \(W_i^u,W_j^v\in \mathbb{R}^d\) are \(d-\)dimensional word embedding vectors. The objective is to predict label \(y\) which indicates whether the reply to the corresponding comment was sarcastic or not.

\subsection{Intra-Sentence Word Encoder Layer}
The primary purpose of this layer is to summarize intra-sentence contextual information from both directions in both the sentences (comment \& reply) using Bidirectional Long Short Term Memory Networks (Bi-LSTM). A Bi-LSTM \cite{650093} processes information in both the directions using a forward LSTM \cite{doi:10.1162/neco.1997.9.8.1735} \(\overrightarrow{h}\), that reads the sentence  \(S = [w_1,w_2,....,w_n]\) from \(w_1\) to \(w_n\) and a backward LSTM \(\overleftarrow{h}\) that reads the sentence from \(w_n\) to \(w_1\). Hidden states from both the LSTMs are added to get the final hidden state representations of each word. So the hidden state representation of the \(t_{th}\) word (\(h_t\)) can be represented by the sum of \(t_{th}\) hidden representations of the forward and backward LSTMs (\(\overrightarrow{h_t}\),\(\overleftarrow{h_t}\)) as show in equations below.
\begin{equation} \label{lstmf}
\overrightarrow{h_t} = \overrightarrow{LSTM}(w_t,\overrightarrow{h_{t-1}}); \overleftarrow{h_t} = \overleftarrow{LSTM}(w_t,\overleftarrow{h_{t-1}})
\end{equation}
\begin{equation} \label{hsum}
h_t = \overleftarrow{h_t}+\overrightarrow{h_t}
\end{equation}

This Intra-Sentence Word Encoder Layer consists of two independent Bidirectional LSTMs for both comment (\(BiLSTM_c\)) and reply (\(BiLSTM_r\)). Apart from the hidden states, both these Bi-LSTMs also generate separate (forward \& backward) final cell states represented by \(\overleftarrow{C}\) \& \(\overrightarrow{C}\). The comment sentence  \(U\) is given as an input to \(BiLSTM_c\) and the reply sentence \(V\) is given as an input to \(BiLSTM_r\). The outputs of both the Bi-LSTMs are represented by the equations \ref{bi1} and \ref{bi2}.
\begin{equation} \label{bi1}
\overrightarrow{C_u},h^u,\overleftarrow{C_u} = BiLSTM_c(U)
\end{equation}
\begin{equation} \label{bi2}
\overrightarrow{C_v},h^v,\overleftarrow{C_v} = BiLSTM_r(V)
\end{equation}

Here, \(\overrightarrow{C_u}, \overrightarrow{C_v} \in \mathbb{R}^d\) are the final cell states of the forward LSTMs corresponding to \(BiLSTM_c\) \& \(BiLSTM_r\); \(\overleftarrow{C_u}, \overleftarrow{C_v} \in \mathbb{R}^d\) are the final cell states of the backward LSTMs corresponding to \(BiLSTM_c\) \& \(BiLSTM_r\); \(h^u = [h_1^u,h_2^u,....,h_n^u]\) and \(h^v = [h_1^v,h_2^v,....,h_n^v]\) are the hidden state representations of \(BiLSTM_c\) \& \(BiLSTM_r\) respectively, where \(h_i^u,h_j^v \in \mathbb{R}^d\) and \(h^u,h^v \in \mathbb{R}^{n \times d}\).

\subsection{Bi-ISCA: Bidirectional Inter-Sentence Contextual Attention Mechanism}
Sarcasm is context-dependent in nature. Even humans sometimes have a hard time understanding sarcasm without having any contextual information. The hidden states generated by both the Bi-LSTMs (\(BiLSTM_c\) \& \(BiLSTM_r\)) captures the intra-sentence bidirectional contextual information in  comment \& reply respectively, but fails to capture the inter-sentence contextual information between them. This paper introduces a novel Bidirectional Inter-Sentence Contextual Attention mechanism (Bi-ISCA) for capturing the inter-sentence contextual information between both the sentences.  

Bi-ISCA uses hidden state representations of \(U\) \& \(V\) along with the auxiliary sentence's cell state representations (\(\overrightarrow{C} \& \overleftarrow{C}\))  to capture the inter-sentence contextual information. At first, the attention mechanism captures four sets of attentions scores namely, (\(\alpha^{\overrightarrow{C_u}},\alpha^{\overleftarrow{C_u}},\alpha^{\overrightarrow{C_v}},\alpha^{\overleftarrow{C_v}} \in \mathbb{R}^n\)). These sets of inter-sentence attention scores are used to generate new inter-sentence contextualized hidden representations. Then (\(\alpha^{\overrightarrow{C_u}},\alpha^{\overleftarrow{C_u}}\)) are calculated using the hidden state representations of \(BiLSTM_r\) along with the forward and backward final states (\(\overrightarrow{C_u} , \overleftarrow{C_u}\)) of \(BiLSTM_c\) (as shown in equations \ref{cu1} \& \ref{cu2}), similarly (\(\alpha^{\overrightarrow{C_v}},\alpha^{\overleftarrow{C_v}}\)) are calculated using the hidden state representations of \(BiLSTM_c\) along with the forward and backward final states (\(\overrightarrow{C_v} , \overleftarrow{C_v}\)) of \(BiLSTM_r\) (as shown in equations \ref{cv1} \& \ref{cv2}). In the equations below (\(\bullet\)) represents a dot product between two vectors.
\begin{equation} \label{cu1}
\alpha^{\overrightarrow{C_u}} = [\alpha_1^{\overrightarrow{C_u}}, \alpha_2^{\overrightarrow{C_u}}, ....,\alpha_n^{\overrightarrow{C_u}}]; \alpha_i^{\overrightarrow{C_u}}=\overrightarrow{C_u} \bullet h_i^v
\end{equation}
\begin{equation} \label{cu2}
\alpha^{\overleftarrow{C_u}} = [\alpha_1^{\overleftarrow{C_u}}, \alpha_2^{\overleftarrow{C_u}}, ....,\alpha_n^{\overleftarrow{C_u}}]; \alpha_i^{\overleftarrow{C_u}}=\overleftarrow{C_u} \bullet h_i^v
\end{equation}
\begin{equation} \label{cv1}
\alpha^{\overrightarrow{C_v}} = [\alpha_1^{\overrightarrow{C_v}}, \alpha_2^{\overrightarrow{C_v}}, ....,\alpha_n^{\overleftarrow{C_v}}]; \alpha_i^{\overrightarrow{C_v}}=\overrightarrow{C_v} \bullet h_i^u
\end{equation}
\begin{equation} \label{cv2}
\alpha^{\overleftarrow{C_v}} = [\alpha_1^{\overleftarrow{C_v}}, \alpha_2^{\overleftarrow{C_v}}, ....,\alpha_n^{\overleftarrow{C_v}}]; \alpha_i^{\overleftarrow{C_v}}=\overleftarrow{C_v} \bullet h_i^u
\end{equation}

In the next step, the above calculated sets of inter-sentence attention scores \(\alpha^{\overrightarrow{C_u}}, \alpha^{\overleftarrow{C_u}}\)) are multiplied back with the hidden state representations of \(BiLSTM_r\) to generate two new set of hidden representations \(h_{v}^{\overrightarrow{C_u}}, h_{v}^{\overleftarrow{C_u}} \in \mathbb{R}^{n \times d}\) of the reply sentence namely, reply contextualized on comment (forward)  \& reply contextualized on comment (backward) respectively (as shown in equations \ref{hv1} \& \ref{hv2}). Similarly \(\alpha^{\overrightarrow{C_v}}, \alpha^{\overleftarrow{C_v}}\) are multiplied back with the hidden state representations of \(BiLSTM_c\) to generate two new set of hidden representations \(h_{u}^{\overrightarrow{C_v}}, h_{u}^{\overleftarrow{C_v}} \in \mathbb{R}^{n \times d}\) of the comment sentence namely, comment contextualized on reply (forward)  \& comment contextualized on reply (backward) respectively (as shown in equations \ref{hu1} \& \ref{hu2}). In the equations below (\(\times\)) represents multiplication between a scalar and a vector.

\begin{equation} \label{hv1}
h_{v}^{\overrightarrow{C_u}} = [h_{v,1}^{\overrightarrow{C_u}},h_{v,2}^{\overrightarrow{C_u}},....,h_{v,n}^{\overrightarrow{C_u}}],; h_{v,i}^{\overrightarrow{C_u}}=\alpha_i^{\overrightarrow{C_u}} \times h_i^v
\end{equation}
\begin{equation} \label{hv2}
h_{v}^{\overleftarrow{C_u}} = [h_{v,1}^{\overleftarrow{C_u}},h_{v,2}^{\overleftarrow{C_u}},....,h_{v,n}^{\overleftarrow{C_u}}],; h_{v,i}^{\overleftarrow{C_u}}=\alpha_i^{\overleftarrow{C_u}} \times h_i^v
\end{equation}
\begin{equation} \label{hu1}
h_{u}^{\overrightarrow{C_v}} = [h_{u,1}^{\overrightarrow{C_v}},h_{u,2}^{\overrightarrow{C_v}},....,h_{u,n}^{\overrightarrow{C_v}}],; h_{u,i}^{\overrightarrow{C_v}}=\alpha_i^{\overrightarrow{C_v}} \times h_i^u
\end{equation}
\begin{equation} \label{hu2}
h_{u}^{\overleftarrow{C_v}} = [h_{u,1}^{\overleftarrow{C_v}},h_{u,2}^{\overleftarrow{C_v}},....,h_{u,n}^{\overleftarrow{C_v}}],; h_{u,i}^{\overleftarrow{C_v}}=\alpha_i^{\overleftarrow{C_v}} \times h_i^u
\end{equation}
\subsection{Integration and Final Prediction}
The proposed model uses Convolutional Neural Networks (CNN) \cite{726791} for capturing location-invariant local features from the newly obtained contextualized hidden representations \(h_{u}^{\overleftarrow{C_v}},h_{u}^{\overrightarrow{C_v}},h_{v}^{\overleftarrow{C_u}},h_{v}^{\overrightarrow{C_u}}\). Four independent CNN blocks (\(CNN_1,CNN_2,CNN_3,CNN_4\)) are used, corresponding to each of the newly obtained contextualized hidden representations. Each \(CNN\) block consists two convolutional layers. Both the convolution layer consist of \(k\) filters of height \(h\). The role of these filters is to detect particular features at different locations of the input. The output \(c_i^l\) of the \(l^{th}\) layer  consists of \(k^l\) feature maps of height \(h\). The \(i^{th}\) feature map (\(c_i^l\)) is calculated as:
\begin{equation} \label{cnn}
c_i^l = b_i^l + \sum_{k^{l-1}}^{j=1}{K_{i,j}^{l} \ast c_j^{l-1}}
\end{equation}

In the above equation, \(b_i^l\) is a bias matrix and \(K_{i,j}^{l}\) is a filter connecting \(j^{th}\) feature map of layer \((l-1)\) to the \(i^{th}\) feature map of layer \((l)\). The output of each convolution layer is passed through a activation function \(f\). The proposed model uses \textit{Leaky}ReLu as its activation function.
\begin{numcases}{f=}
    a \ast x, & for $x \geq 0; a \in \mathbb{R} $\\
    x, & for $x < 0$
\end{numcases}

For each of the CNN blocks, the corresponding contextualized hidden representations are first concatenated (\(\oplus\)) and then given as input. The outputs of all the CNN blocks are flattened (\(F_1,F_2,F_3,F_4 \in \mathbb{R}^{dk}\)) and concatenated to generate a new vector (\(p\in \mathbb{R}^{4dk}\)), where \(d\) represents the dimension of the hidden representation and \(k\) represents number of convolutional filters used. This concatenated (\(p\)) vector is then given as input to a dense layer having \(4dk\) neurons and is followed by the final sigmoid prediction layer.  
\begin{equation} \label{f1}
F_1 = CNN_1([h_{u,1}^{\overrightarrow{C_v}} \oplus h_{u,2}^{\overrightarrow{C_v}} \oplus .... \oplus h_{u,n}^{\overrightarrow{C_v}}])
\end{equation}
\begin{equation} \label{f2}
F_2 = CNN_2([h_{u,1}^{\overleftarrow{C_v}} \oplus h_{u,2}^{\overleftarrow{C_v}} \oplus .... \oplus h_{u,n}^{\overleftarrow{C_v}}])
\end{equation}
\begin{equation} \label{f3}
F_3 = CNN_3([h_{v,1}^{\overrightarrow{C_u}} \oplus h_{v,2}^{\overrightarrow{C_u}} \oplus .... \oplus h_{v,n}^{\overrightarrow{C_u}}])
\end{equation}
\begin{equation} \label{f4}
F_4 = CNN_4([h_{v,1}^{\overleftarrow{C_u}} \oplus h_{v,2}^{\overleftarrow{C_u}} \oplus .... \oplus h_{v,n}^{\overleftarrow{C_u}}])
\end{equation}
\begin{equation} \label{c}
p = [F_1 \oplus F_2 \oplus F_3 \oplus F_4]
\end{equation}
\begin{equation} \label{sig}
\hat y = \sigma(W p + b), \quad W \in \mathbb{R}^{4dk};b \in \mathbb{R} 
\end{equation}

The proposed model uses the binary cross-entropy as the training loss function as shown in equation \ref{loss}. Here (\(L\)) is the cost function, \(\hat y_i \in \mathbb{R}\) represents the output of the proposed model, \(y_i \in \mathbb{R}\) represents the true label and \(N \in \mathbb{N}\) represents the number of training samples. 
\begin{equation} \label{loss}
L=-\frac{1}{N}\sum_{i=1}^{N}{y_i \cdot log(\hat y_i) + (1-y_i) \cdot log(1-\hat y_i)}  
\end{equation}

\section{Evaluation Setup} \label{evaluation}

\subsection{Dataset}
This paper focuses on detecting sarcasm in the user-generated short text using only the conversational context. Social media platforms like Reddit and Twitter are widely used by users for posting opinions and replying to other's opinions. They have proved to be of a great source for extracting conversational data. So the experiments were conducted on two publicly available benchmark datasets (Reddit \& Twitter) used for the sarcasm detection task. Both the datasets consist of comments and reply pairs.

\begin{table}
\caption{Statics of the SARC dataset and FigLang 2020 workshop Twitter dataset.}
\label{tab:datasets}
\centering
\resizebox{\textwidth}{!}{%
\begin{tabular}{lll|cc|cc|cc}
\hlineB{3}
\multicolumn{3}{l}{\multirow{2}{*}{}} & \multicolumn{2}{|c|}{No. of comment-reply pairs} & \multicolumn{2}{c|}{Avg. no. of words per comment} & \multicolumn{2}{c}{Avg. no. of words per reply} \\ \cline{4-9} 
\multicolumn{3}{l}{} & \multicolumn{1}{|c|}{Sarcastic} & Non-Sarcastic & \multicolumn{1}{c|}{Sarcastic} & Non-Sarcastic & \multicolumn{1}{c|}{Sarcastic} & Non-Sarcastic \\ \hlineB{3}
\multicolumn{1}{l|}{\multirow{3}{*}{Training set}} & \multicolumn{1}{l|}{\multirow{2}{*}{Reddit}} & Balanced & 81205 & 81205 & 12.69 & 12.67 & 12.19 & 12.21 \\ \cline{3-3}
\multicolumn{1}{l|}{} & \multicolumn{1}{l|}{} & Imbalanced & 16303 & 81205 & 12.69 & 12.65 & 12.15 & 12.21 \\ \cline{2-3}
\multicolumn{1}{l|}{} & \multicolumn{1}{l|}{Twitter} & Balanced & 3496 & 3496 & 24.97 & 24.97 & 24.25 & 24.25 \\ \hline \hline
\multicolumn{1}{l|}{\multirow{3}{*}{Testing set}} & \multicolumn{1}{l|}{\multirow{2}{*}{Reddit}} & Balanced & 9058 & 9058 & 12.71 & 12.64 & 12.14 & 12.22 \\ \cline{3-3}
\multicolumn{1}{l|}{} & \multicolumn{1}{l|}{} & Imbalanced & 1747 & 9058 & 12.73 & 12.69 & 12.20 & 12.21 \\ \cline{2-3}
\multicolumn{1}{l|}{} & \multicolumn{1}{l|}{Twitter} & Balanced & 874 & 874 & 24.97 & 24.97 & 24.25 & 24.25 \\ \hlineB{3}
\end{tabular}%
}
\end{table}

\textbf{SARC\footnote{https://nlp.cs.princeton.edu/SARC/2.0/} Reddit \citep{khodak2018corpus}} is the largest dataset available for sarcasm detection containing millions of sarcastic/non-sarcastic comments-reply pairs from the social media site Reddit. This dataset was generated by scraping comments from Reddit containing the \textbackslash s (sarcasm) tag. It contains replies, their parent comment (acts as context), and a label that shows whether the reply was sarcastic/non-sarcastic to their corresponding parent comment. To compare the performance of the model on a different dataset (latest), the proposed model was also evaluated on the Twitter dataset provided in the \textbf{FigLang\footnote{sites.google.com/view/figlang2020} 2020 workshop \citep{ghosh-etal-2020-report}} for the "sarcasm detection shared task". This consists of sarcastic/non-sarcastic tweets and their corresponding contextual parent tweets. The sarcastic tweets were collected using hashtags like \#sarcasm, \#sarcastic, and \#irony, similarly non-sarcastic tweets were collected using hashtags like \#happy, \#sad, and \#hate. This dataset sometime contains more than one contextual parent tweet, so in those cases, all of the contextual tweets are considered independently with the target tweet. 

In both the datasets, replies are the target comment/tweet to be classified as sarcastic/non-sarcastic, and their corresponding parent comment/tweet acts as context. Both the datasets constitute of comments/tweets of varying lengths, but because this paper only focuses on detecting sarcasm in the short text, only the short comment/reply pairs were used. Comment/reply sentences of length (no. of words) less than 20, 40 were used in the case of SARC and Twitter dataset respectively. In both cases, the balanced datasets contain equal proportions of sarcastic/non-sarcastic comment/reply pairs, and the imbalanced datasets maintain a 20:80 ratio (approximately) between sarcastic and non-sarcastic comment/reply pairs. Testing was done on 10\% of the dataset and the rest was used for training. 10\% of the training set was used for validation purposes. Statistics of both the datasets are shown in Table \ref{tab:datasets}.

\subsection{Data Preprocessing}
The preprocessing of the textual data was done by first lower-casing all the sentences and separating punctuations from the words. We do not remove the stop-words because we believe that sometimes stop-words play a major role in making a sentence sarcastic e.g., \textit{"is it?"} and \textit{"am I?"}. The problem with social media platforms is that, users use a lot of abbreviations, shortened words and slang words like, \textit{"IMO"} for \textit{"in my opinion"}, \textit{lmk"} for \textit{"let me know "}, \textit{"fr"} for \textit{"for"}, etc. These words are challenging to taken care of in the NLP tasks, particularly in the automatic discovery of flexible word usages. So to solve this problem, these words are converted to their corresponding full-forms using abbreviation/slang word dictionaries obtained from urban dictionary\footnote{https://www.urbandictionary.com/}. After this, all the sentences were tokenized into a list of words. The proposed model had a fixed input size for both comment and reply, but not all the sentences were of the same length. So all the sentences were padded to the length of the longest sentence (20 in the case of the Reddit dataset and 40 in the case of the Twitter dataset). Word embeddings are used to give semantically-meaningful dense representations to the words. Word-based embeddings are constructed using contextual words whereas character-based embeddings are constructed from character n-grams of the words.  Character-based in contrast to the Word-based embeddings solves the problem of out of vocabulary words and performs better in the case of infrequent words by creating word embeddings based only on their spellings. So for generating proper representations for words we have used FastText\footnote{https://fasttext.cc/}, a character-based word embedding. This would not only give words better representation compared to the word-based model but also incorporate slang/shortened/infrequent words (which commonly appear in social media platforms).

\subsection{Training Details}
We have used macro-averaged (\textit{F1}) and accuracy (\textit{Acc}) scores as the evaluation metric, as it is standard for the sarcasm detection task. We have also reported Precision (\textit{P}) and Recall (\textit{R}) scores in the case of the Twitter dataset as well as for the Reddit dataset (wherever available). Hyperparameter tuning was used to find optimum values of the hyperparameters. The FastText embeddings used were of size \(d=30\) and were trained for 30 iterations having window size of 3, 5 in the case of SARC, and Twitter dataset respectively. The number of filters in all the convolutional blocks were [64, 64] of height [2, 2]. The learning optimizer used is Adam with an initial learning rate of 0.01. The value of $\alpha$ in all the \textit{Leaky}ReLu layers was set to 0.3. All the models were trained for 20 epochs. L2 regularization set to \(10^{-2}\) is applied to all the feed-forward connections along with early stopping having the patience of 5 to avoid overfitting. The mini-batch size was tuned amongst \{100, 500, 1000, 2000, 3000, 4000\} and was observed that mini-batch size of 2000, 500 gave the best performance for the SARC and Twitter dataset respectively.

The recent success of transformer-based language models has led to their wide usage in sentiment analysis tasks. They are known for generating high quality high dimensional word representations (768-dimensional for BERT). Their only drawback is that they require high processing power and memory to train. The above-mentioned configuration of the proposed model generates $\approx$1120K trainable parameters, and increasing either the embedding size or the number of tokens in a sentence led to an exponential increase in the number of trainable parameters. So due to computational resource limitations, we limited our experiments to lower-dimensional word embeddings.

\section{Results} \label{result}

Bi-ISCA focuses on only using the contextual comment/tweet for detecting sarcasm rather than using any other topical/personality-based features. Using only the contextual information enriches the model's ability to capture syntactical and semantical textual properties responsible for invoking sarcasm in any type of conversation. Table \ref{tab:sarc-result} reports performance results on the SARC  datasets. For comparison purposes, F1-score (\textit{F1}), Accuracy score (\textit{Acc}), Precision (\textit{P}) and Recall (\textit{R}) were used.
\begin{table}[t]
    \caption{(a) Results on the SARC dataset. Models with \colorbox{Blue}{blue} backgrounds use only contextual text for detecting sarcasm. (b) Results on the FigLang 2020 workshop Twitter dataset.}%
    \centering
    \subfloat[]{
    {\resizebox{0.5\textwidth}{!}{%
    \begin{tabular}{l|cccc|cccc}
    \hlineB{3.5}
    
    \multirow{2}{*}{Models} & \multicolumn{4}{c|}{Balance}                    & \multicolumn{4}{c}{Imbalanced}                 \\ \cline{2-9} 
                                      & \multicolumn{1}{c|}{\textit{Acc}} & \multicolumn{1}{c|}{\textit{F1}} & \multicolumn{1}{c|}{\textit{P}} & \multicolumn{1}{c|}{\textit{R}} & \multicolumn{1}{c|}{\textit{Acc}} & \multicolumn{1}{c|}{\textit{F1}} & \multicolumn{1}{c|}{\textit{P}} & \multicolumn{1}{c}{\textit{R}} \\ \hlineB{1} \hlineB{1}
    CNN-SVM \citep{DBLP:journals/corr/PoriaCHV16} $^{\textcolor{blue}{\dagger}\textcolor{purple}{ \star  }}$& 68.0 & 68.0 & -- & -- & 69.0 & 79.0 & -- & -- \\
    \rowcolor{Blue} AMR \citep{DBLP:journals/corr/abs-1809-03051} $^{\textcolor{red}{\ddagger}}$   & 69.5 & 69.5 & 74.8 & 69.7 & -- & -- & -- & -- \\
    \rowcolor{Blue} \citep{DBLP:journals/corr/GhoshFM17} $^{\textcolor{red}{\ddagger}}$ & \multirow{1}{*}{--} & \multirow{1}{*}{67.8} & \multirow{1}{*}{68.2} & \multirow{1}{*}{67.9}  & \multirow{1}{*}{--} & \multirow{1}{*}{--} & \multirow{1}{*}{--} & \multirow{1}{*}{--} \\
    CUE-CNN \citep{DBLP:journals/corr/AmirWLCS16} $^{\textcolor{blue}{\dagger}\textcolor{purple}{ \star  }}$& 70.0 & 69.0 & -- & -- & 73.0 & 81.0 & -- & -- \\
    MHA-BiLSTM \citep{8949523} $^{\textcolor{blue}{\dagger}}$ & -- & 77.5 & 72.6 & 83.0 & -- & 56.8 & 60.3 & 53.7 \\ \hlineB{1.5}
    CASCADE \citep{hazarika-etal-2018-cascade} $^{\textcolor{red}{\ddagger}\textcolor{purple}{ \star  }}$& 77.0 & 77.0 & -- & -- & 79.0 & 86.0 & -- & -- \\
    \rowcolor{Blue} CASCADE (only discourse features) $^{\textcolor{red}{\ddagger}}$  & 68.0 & 66.0 & -- & -- & 68.0 & 78.0 & -- & -- \\ \hlineB{1} \hlineB{1}
    \rowcolor{Blued} \textbf{Bi-ISCA (this paper) $^{\textcolor{red}{\ddagger}}$}     & \textbf{72.3}  & \textbf{75.7} & \textbf{74.2} & \textbf{77.6} & \textbf{71.9} & \textbf{74.4} & \textbf{73.0} & \textbf{75.8} \\ \hlineB{3}
    \% increase w.r.t CASCADE & \multirow{2}{*}{\textit{\textcolor{agr}{$6.3 \uparrow $}}} & \multirow{2}{*}{\textit{\textcolor{agr}{$14.7 \uparrow $}}} & -- & -- & \multirow{2}{*}{\textit{\textcolor{agr}{$5.7 \uparrow $}}} & \multirow{2}{*}{\textit{\textcolor{magenta}{$4.6 \downarrow $}}} & -- & -- \\
    (only discourse features) & & & & & & & & \\
    \hlineB{3}
    \multicolumn{5}{l}{ \begin{tabular}[c]{@{}l@{}}\textcolor{blue}{$\ddagger$} Uses only target sentence, \textcolor{red}{$\ddagger$} Uses context along with target sentence,\\ \textcolor{purple}{$ \star $} Uses personality-based features\end{tabular}} 
    
    \end{tabular}
    }}\label{tab:sarc-result}
    }%
    \qquad
    \subfloat[]{
    {\resizebox{0.4\textwidth}{!}{%
    \begin{tabular}{l|c|c|c}
    \hlineB{3}
    \multirow{1}{*}{Models}                          & P             & R             & F1            \\ \hlineB{1.5}
    Baseline (\(LSTM_attn\))                                    & 70.0          & 66.9          & 68.0          \\ 
    BERT-Large+BiLSTM+SVM \citep{baruah-etal-2020-context}                       & 73.4          & 73.5          & 73.4          \\ 
    BERT+CNN+LSTM \citep{srivastava-etal-2020-novel}                               & 74.2          & 74.6          & 74.1          \\ 
    RoBERTa+LSTM \citep{kumar-anand-2020-transformers}                               & 77.3          & 77.4          & 77.2          \\ 
    RoBERT-Large \citep{dong-etal-2020-transformer}                               & 79.1          & 79.4          & 79.0          \\ 
    \begin{tabular}[c]{@{}l@{}}RoBERT+Multi-Initialization Ensemble \\ \citep{jaiswal-2020-neural} \end{tabular}       & 79.2          & 79.3          & 79.1          \\ 
    \begin{tabular}[c]{@{}l@{}}BERT + BiLSTM + NeXtVLAD + Context Ensemble \\ + Data Augmentation \citep{lee-etal-2020-augmenting} \end{tabular}    & 93.2          & 93.6          & 93.1          \\ \hlineB{1.5}
    \textbf{Bi-ISCA (this paper)}               & \textbf{89.4} & \textbf{94.8} & \textbf{91.7} \\ \hlineB{2}
    \end{tabular}
    }}\label{tab:twitter-result}
    }%
\end{table}

When compared with the existing works, Bi-ISCA was able to outperform all the models that use only conversational context for sarcasm detection (improvement of 11.7\% in F1 score when compared to \citep{DBLP:journals/corr/GhoshFM17}; 8.9\% in F1 score and 4\% in accuracy when compared to AMR \citep{DBLP:journals/corr/abs-1809-03051}) and was even able to perform better than the models that use personality-based features along with the target sentence for detecting sarcasm (improvement of 11.3\% in F1 and 6.3\% in accuracy score when compared to CNN-SVM \citep{DBLP:journals/corr/PoriaCHV16}; 9.7\% in F1 score and 4.8\% in accuracy when compared to CUE-CNN \citep{DBLP:journals/corr/AmirWLCS16}). MHA-BiLSTM \citep{8949523} had a 2.3\% higher F1 score in the balanced dataset but Bi-ISCA was able to show drastic improvement of 31\% in the imbalanced dataset, which demonstrated the ability of Bi-ISCA to handle class imbalance.

The current state-of-the-art on the SARC dataset is achieved by CASCADE. Even though CASCADE uses personality-based features and contextual information along with large sentences of average length $\approx$55-62 (very large compared to our dataset, which gives them the advantage of using a lot more contextual information), Bi-ISCA was able to achieve an F1 score comparable to it (despite using relatively short text). In comparison with CASCADE that only uses discourse-based features, Bi-ISCA performed drastically better with an increase of 14.7\% in F1 and 6.3\% in accuracy score for the balanced dataset. 

Bi-ISCA clearly demonstrated its capabilities to robustly handle an imbalance in the dataset, although it was unable to outperform both the CASCADE models. This slightly poor performance in the imbalanced dataset can be explained by the length of sentences used by CASCADE, which are significantly ($\approx$5 times) greater than the ones on which Bi-ISCA was tested. Longer sentences result in increased contextual information which improves performance especially in the case of imbalance where little extra information can lead to a drastic increase in performance. 

Table \ref{tab:twitter-result} reports Precision (\textit{P}), Recall (\textit{R}), and F1-score (\textit{F1}) of different models from the leaderboard of FigLang 2020 sarcasm detection shared task using the Twitter dataset. In this case, not only Bi-ISCA was able to outperform the baseline model \citep{ghosh-etal-2020-report} (improvement of 19\%, 28\% \& 24\% in precision, recall, and F1 score respectively), but was also able to perform comparably to the state-of-the-art \citep{lee-etal-2020-augmenting} with a 1.2\% increase in recall, which further validates the performance of the proposed model. Even though all the models other than the baseline in Table \ref{tab:twitter-result} are a transformer-based model, Bi-ISCA was able to outperform them all.

\section{Discussion} \label{discussion}
\begin{table}[t]
\caption{Attension weight distribution in reddit comment-reply pairs. Here \textit{CcR} represents "Comment contextualized on Reply" whereas \textit{RcC} represents "Reply contextualized on Comment"; \textit{\textbf{(R)}} \& \textit{\textbf{(L)}} represents forward \& backward attention.}
\label{tab:att_ex}
\centering
\resizebox{0.65\textwidth}{!}{%
\begin{tabular}{r|c|l}
\hlineB{3}
1. & \includegraphics[width=0.51in]{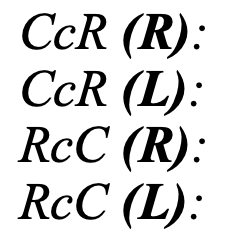} &  \includegraphics[width=3.4in]{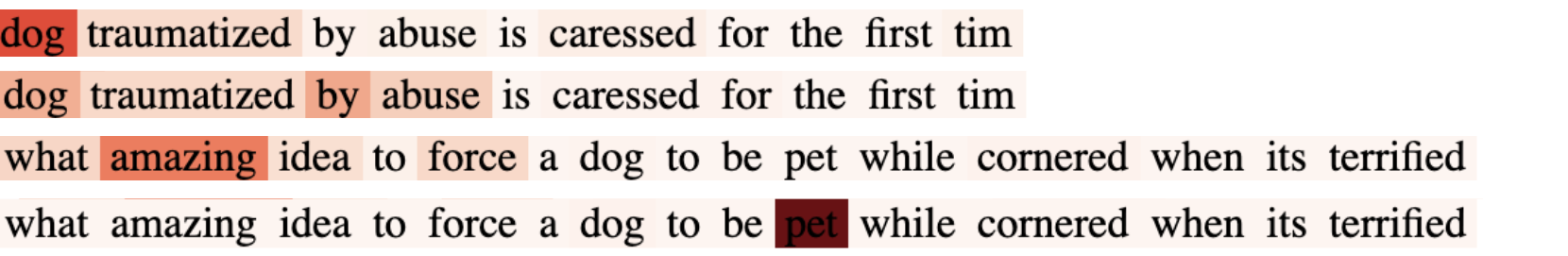} \\ \hline
2. & \includegraphics[width=0.51in]{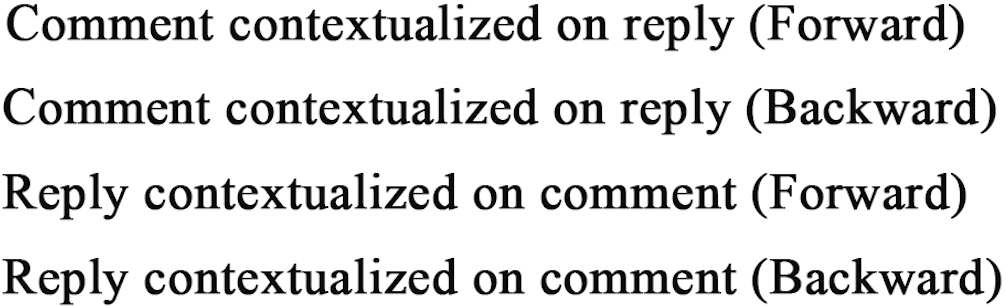} &  \includegraphics[width=3.6in]{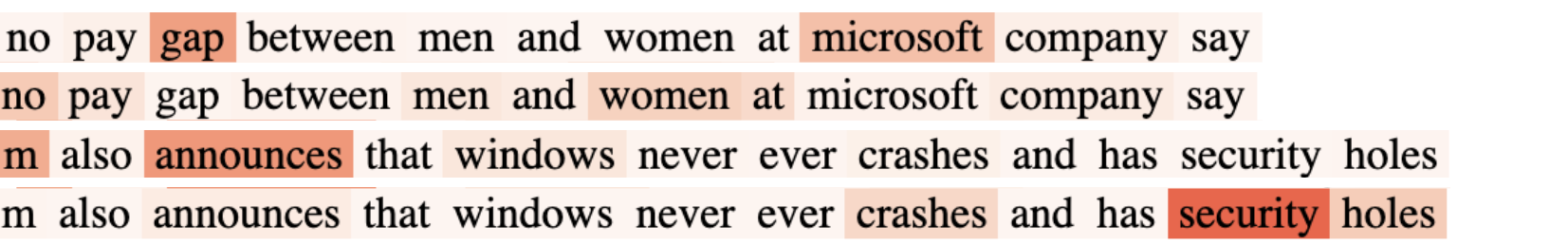} \\ \hline 
3. & \includegraphics[width=0.51in]{sentences/main.png} &  \includegraphics[width=3.6in]{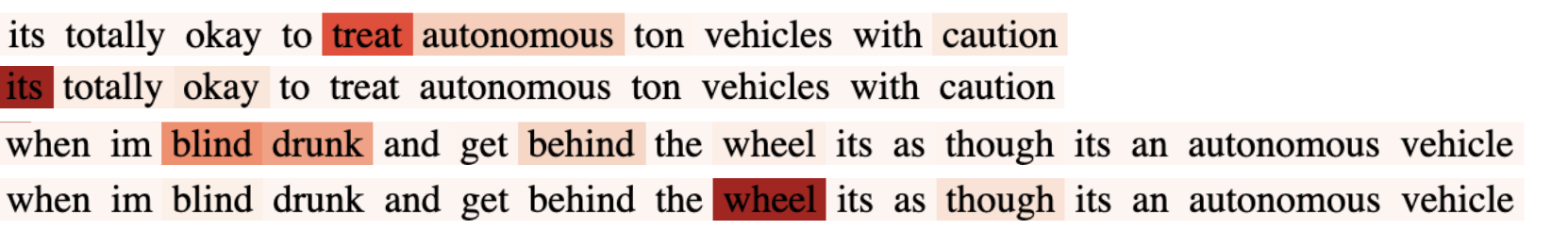} \\ \hline 
4. & \includegraphics[width=0.51in]{sentences/main.png} &  \includegraphics[width=3.4in]{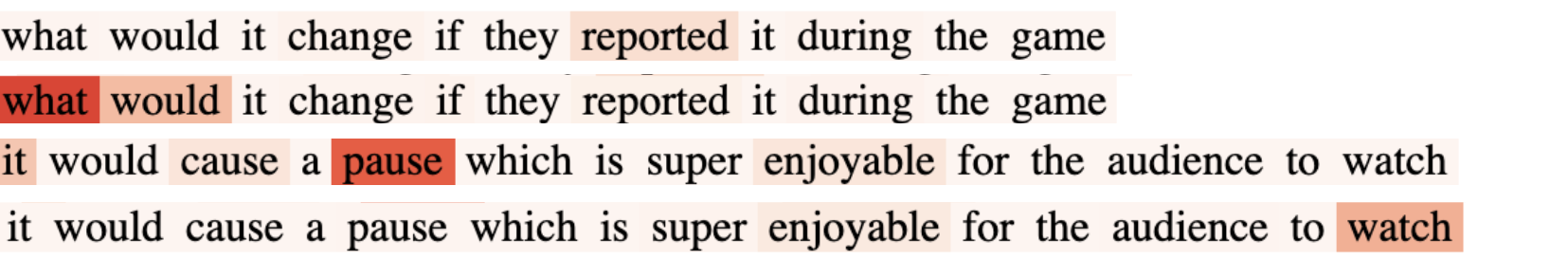} \\ \hline
\end{tabular}
}
\end{table}
The attention scores generated by the attention mechanism makes the proposed model highly interpretable. Table \ref{tab:att_ex} showcases the distribution of the attention scores over four sarcastic (correctly predicted by Bi-ISCA) comment-reply pairs from the SARC dataset. Not only the proposed model was correctly able to detect sarcasm in these pairs of sentences but was also able to correctly identify words responsible for contextual, explicit, or implicit incongruity which invokes sarcasm. 

For example in Pair 1, Bi-ISCA correctly identified explicitly incongruous words like \textit{"amazing"} and \textit{"force"} in the reply sentence which were responsible for the sarcastic nature of the reply. Interestingly the word "traumatized" in the parent comment also had a high attention weight value, which shows that the proposed attention mechanism was able to learn the contextual incongruity between the opposite sentiment words like "traumatized" \& "amazing" in the comment-reply pair. Pair 2 demonstrates the model's ability to capture words responsible for invoking sarcasm by making sentences implicitly incongruous. Sarcasm due to implicit incongruity is usually the toughest to perceive. Despite this, Bi-ISCA was able to give high attention weights to words like "announces" and "crashes \& security holes". Not only this, but the proposed intra-sentence attention mechanism was also able to learn a link between "microsoft" and "m" (slang for microsoft) without having any prior knowledge related to slangs. Pair 3 is also an example of an explicitly and contextually incongruous comment-reply pair, where the model was successfully able to capture opposite sentiment words \& phrases like "blind drunk", "cautious" and "behind the wheel" that made the reply sarcastic in nature. Pair 4 is an example of sarcasm due to implicit incongruity between the words, "pause" \& "watch", and contextual incongruity simultaneously between "reported" \& "enjoyable", both of which were successfully captured by Bi-BISCA.

\section{Conclusion} \label{conclude}
In this paper, we introduced STEPs-RL for learning phonetically sound spoken-word representations using speech and text entanglement. Our approach achieved an accuracy of 89.47\% in predicting phonetic sequences when both gender and dialect of the speaker are used in the auxiliary information. We also compared its performance using different configurations and observed that the performance of the proposed model improved by (1) increasing the spoken word latent representation size, and (2) the addition of auxiliary information like gender and dialect. We were not only able to validate the capability of the learned representations to capture the semantical and syntactical relationships between the spoken-words but were also able to illustrate soundness in the phonetic structure of the generated vector space. For future work, we plan to (1) extend the model to use attention mechanisms, (2) improve performance by using transformer-based architecture, and (3) experimenting on larger datasets, and (4) using features other than MFCCs.

\bibliography{neurips_2020}
\bibliographystyle{acl_natbib}

\end{document}